%% file: main.tex
\documentclass[preprint,review,11pt]{elsarticle}
\usepackage[top=3.0cm,left=2.4cm,right=2.4cm,bottom=3.0cm]{geometry}

\usepackage{amssymb}
\usepackage{graphicx}
\usepackage{subcaption}
\usepackage{amsmath}
\usepackage{comment}
\usepackage{amssymb}
\usepackage{booktabs}
\usepackage{tabularray}
\usepackage{bm}
\usepackage{dsfont}
\usepackage{booktabs}
\usepackage[referable]{threeparttablex}
\usepackage{multirow}
\usepackage{numprint}
\usepackage[pagebackref,breaklinks,colorlinks]{hyperref}

\usepackage[capitalize]{cleveref}
\crefname{section}{Sec.}{Secs.}
\Crefname{section}{Section}{Sections}
\Crefname{table}{Table}{Tables}
\crefname{table}{Tab.}{Tabs.}

\journal{Engineering Applications of Artificial Intelligence}

\hypersetup{pdfauthor={Luca Della Libera, Jacopo Andreoli, Davide Dalle Pezze, Mirco Ravanelli, Gian Antonio Susto},
pdftitle={Bayesian Deep Learning for Remaining Useful Life Estimation via Stein Variational Gradient Descent},
pdfsubject={Remaining useful life estimation},
pdfkeywords={Prognostics and health management, Remaining useful life, Bayesian deep learning, Stein variational gradient descent}}

\begin{document}

\begin{frontmatter}


\title{Bayesian Deep Learning for Remaining Useful Life Estimation via Stein Variational Gradient Descent}


\author[inst1]{Luca Della Libera\corref{firstcorr}}
\cortext[firstcorr]{Corresponding author.}
\ead{luca.dellalibera@mail.concordia.ca}

\affiliation[inst1]{organization={Concordia University, Gina Cody School of Engineering and Computer Science},%Department and Organization
            addressline={1455 Boul. de Maisonneuve Ouest}, 
            city={Montreal},
            postcode={H3G 1M8}, 
            state={Quebec},
            country={Canada}}

\author[inst2]{Jacopo Andreoli}
\ead{jacopo.andreoli@dei.unipd.it}

\author[inst2]{Davide Dalle Pezze}
\ead{davide.dallepezze@phd.unipd.it}

\author[inst1,inst3]{Mirco Ravanelli}
\ead{mirco.ravanelli@concordia.ca}

\author[inst2]{Gian Antonio Susto}
\ead{gianantonio.susto@unipd.it}

\affiliation[inst2]{organization={Università degli Studi di Padova, Dipartimento di Ingegneria dell'Informazione},%Department and Organization
            addressline={Via Gradenigo 6/b}, 
            city={Padova},
            postcode={35131}, 
            state={Veneto},
            country={Italy}}

\affiliation[inst3]{organization={Mila-Quebec AI Institute},%Department and Organization
            addressline={6666 Rue Saint-Urbain}, 
            city={Montreal},
            postcode={H2S 3H1}, 
            state={Quebec},
            country={Canada}}

\begin{abstract}
A crucial task in predictive maintenance is estimating the remaining useful life of physical systems.
In the last decade, deep learning has improved considerably upon traditional model\nobreakdash-based and statistical approaches in terms of predictive performance. However, in order to optimally plan maintenance operations, it is also important to quantify the uncertainty inherent to the predictions. This issue can be addressed by turning standard frequentist neural networks into Bayesian neural networks, which are naturally capable of providing confidence intervals around the estimates. Several methods exist for training those models. Researchers have focused mostly on parametric variational inference and sampling\nobreakdash-based techniques, which notoriously suffer from limited approximation power and large computational burden, respectively. In this work, we use Stein variational gradient descent, a recently proposed algorithm for approximating intractable distributions that overcomes the drawbacks of the aforementioned techniques.
In particular, we show through experimental studies on simulated run\nobreakdash-to\nobreakdash-failure turbofan engine degradation data that Bayesian deep learning models trained via Stein variational gradient descent consistently outperform with respect to convergence speed and predictive performance both the same models trained via parametric variational inference and their frequentist counterparts trained via backpropagation. Furthermore, we propose a method to enhance performance based on the uncertainty information provided by the Bayesian models. We release the source code at \href{https://github.com/lucadellalib/bdl-rul-svgd}{https://github.com/lucadellalib/bdl-rul-svgd}.
\end{abstract}


% Should be in a separate editable file
%\item Stein variational gradient descent is used to train Bayesian neural networks for remaining useful life estimation.
%\item A comparative study is performed on simulated turbofan engine run-to-failure data.
%\item The proposed method is shown to be more accurate and to converge faster than other popular techniques.
%\item A novel heuristic is proposed for improving performance based on uncertainty information.


\end{frontmatter}

%% \linenumberset


%------------------------------------------------------------------------
\section{Introduction}
\label{sec:introduction}
Predictive maintenance is a strategy that optimizes maintenance activities based on real\nobreakdash-time monitoring of machinery health conditions. It has become more and more successful in recent years as a result of its efficacy in eliminating needless interventions and enhancing reliability of equipment~\cite{lei_systematic_review}. One of the key tasks in predictive maintenance is estimating the remaining useful life (RUL) of physical systems, i.e. the available time before a failure occurs in one of the system's components. RUL prediction has traditionally been performed using model\nobreakdash-based and statistical methods~\cite{jardine_approaches}. While the former characterize the deterioration process by means of a mathematical model of the underlying failure mechanism, the latter use existing failure data to fit a statistical model without depending on any physical principle.
Artificial intelligence techniques have lately experienced a surge in popularity thanks to their ability to learn deterioration patterns directly from observations.
Deep learning in particular has emerged as a powerful approach for RUL estimation, outperforming traditional prognostic algorithms~\cite{lorenti2023predictive}.
It has gained extensive traction and adoption across a diverse array of domains such as aviation~\cite{tian_dense, gugulothu_rnn, zheng_lstm, rul_cnn_first, rul_dcnn, jayasinghe_tcnn, yu_transformer, liu_transformer, yoon_vae, ellefsen_ga}, semiconductor manufacturing~\cite{susto2016dealing, lorenti2023predictive}, transport~\cite{fink2015classification}, and machine tools~\cite{susto2016dealing}.
Although deep learning models can offer reasonably precise estimates, achieving $100\%$ accuracy is unlikely. Moreover, errors in RUL tasks have completely different impact: for example, overestimating the RUL could lead to unexpected breaks, while underestimating it could result in unnecessary maintenance~\cite{susto2015machine}. Users may want to tune the performances of predictive maintenance tools based on various factors, which can have different importance over time, such the availability of operators and spare parts, the presence of maintenance plans associated with other tasks in the line, production plans, and deadlines associated with product delivery. Thus, in order to make informed decisions on how to schedule maintenance operations, it is critical to quantify the uncertainty inherent to the predictions.
A standard way to achieve that is to adopt a Bayesian viewpoint and treat the model's parameters as random variables, whose prior distribution is updated into a posterior distribution according to Bayes' rule after observing the data.
Unfortunately, exactly computing the posterior is often infeasible when dealing with deep learning models, thus various approximate Bayesian inference schemes such as {Bayes by Backprop}~\cite{weight_uncertainty}, {Monte Carlo dropout}~\cite{mc_dropout}, and {Markov chain Monte Carlo}~\cite{neal_mcmc} have been employed, with promising results in many practical problems, including RUL estimation~\cite{kraus_bayesian, peng_bayesian, huang_bayesian, caceres_bayesian, li_bayesian, benker_bayesian}.
However those techniques present limitations: either they constrain inference to a fixed family of distributions (multivariate normal with diagonal covariance matrix in Bayes by Backprop and Bernoulli in Monte Carlo dropout), or they are computationally intensive and therefore not scalable to big data (Markov chain Monte Carlo).

{Stein variational gradient descent}~\cite{liu_svgd} is a recent gradient\nobreakdash-based variational inference algorithm for approximating intractable distributions that overcomes the drawbacks of the aforementioned approaches. In particular, differently from Monte Carlo dropout and Bayes by Backprop, it is more expressive, as it does not constrain the posterior to a fixed family of distributions, and differently from Markov chain Monte Carlo, it is more efficient, as it is amenable to batch optimization and parallelization. Furthermore, being sampling\nobreakdash-free and easily adaptable to the geometry of the target space, it is more stable and converges faster.
Despite its success in generative modeling~\cite{pu_svae} and deep reinforcement learning~\cite{liu_svpg}, it is still underexplored, and, to the best of our knowledge, it has not been applied to RUL estimation yet.
As a novel contribution to the field, we show through an experimental study on simulated run\nobreakdash-to\nobreakdash-failure turbofan engine degradation data that Bayesian deep learning models trained via Stein variational gradient descent consistently outperform in terms of convergence speed and predictive performance both the same models trained via Bayes by Backprop, which is the de\nobreakdash-facto standard for training large scale Bayesian neural networks~\cite{jospin_tutorial}, and their frequentist (i.e. non\nobreakdash-Bayesian) counterparts trained via backpropagation.
Additionally, we propose a refreshingly simple yet effective heuristic to enhance the performance based on the uncertainty information provided by the Bayesian models.
To foster reproducibility and promote further research in the field, we release the source code at \href{https://github.com/lucadellalib/bdl-rul-svgd}{https://github.com/lucadellalib/bdl-rul-svgd}.

The remainder of this paper is structured as follows. In \cref{sec:related_work} we summarize related work in frequentist and Bayesian deep learning for RUL prediction. In \cref{sec:methodology} we introduce our method. In \cref{sec:experimental_setup} we describe the experimental setup. In \cref{sec:experimental_results_and_discussion} we present and discuss the experimental results. Lastly, in \cref{sec:conclusions_and_future_work} we draw the conclusions and outline future work directions.

%------------------------------------------------------------------------
\section{Related Work}
\label{sec:related_work}
In the last decade, considerable effort has been dedicated to advance the state\nobreakdash-of\nobreakdash-the\nobreakdash-art of RUL estimation through deep learning.
One of the earliest successful works is the one by Tian~\cite{tian_dense}, who developed a dense neural network model to predict the RUL of a group of pumps based on historical vibration data collected from bearings. Despite achieving a relatively low error rate, extensive preprocessing was necessary to extract relevant information from the data due to the weak inductive bias of dense layers, which complicates the learning of temporal and spatial correlations.
To address the issue of modeling temporal correlations, Gugulothu et al.~\cite{gugulothu_rnn} utilized recurrent neural networks for better capturing the deterioration patterns in the multivariate trajectories, obtaining state\nobreakdash-of\nobreakdash-the\nobreakdash-art performance on the turbofan engine degradation dataset from the IEEE PHM 2008 data challenge~\cite{saxena_phm08}. Zheng et al.~\cite{zheng_lstm} later improved upon Gugulothu et al.'s method through long short\nobreakdash-term memory networks~\cite{hochreiter_lstm}.
Other researchers focused instead on extracting spatial correlations between different sensor measurements by means of convolutional neural networks.
Babu et al.~\cite{rul_cnn_first} introduced a novel convolutional neural network architecture for RUL prediction that outperformed a dense neural network, a support vector machine, and a relevance vector machine on NASA's Commercial Modular Aero\nobreakdash-Propulsion System Simulation (C\nobreakdash-MAPSS) dataset~\cite{saxena_cmapss, damage_propagation}. Li et al.~\cite{rul_dcnn} further reduced the error rate compared to both Zheng et al. and Babu et al. with a deeper convolutional neural network architecture regularized via {dropout}~\cite{dropout} and trained through the adaptive learning rate optimizer {Adam}~\cite{adam}.
An attempt to combine the best of both recurrent neural networks and convolutional neural networks was carried out by Jayasinghe et al.~\cite{jayasinghe_tcnn}, with promising results for data obtained from complex environments.
More recently, transformer\nobreakdash-based methods~\cite{vaswani_transformer}, widely used in natural language processing and computer vision, have been employed in predictive maintenance too with excellent results~\cite{yu_transformer, liu_transformer, dalle2023multi}.
Other techniques such as variational autoencoders~\cite{kingma_vae} and restricted Boltzmann machines~\cite{ackley_rbm} have been applied as well in a semi\nobreakdash-supervised fashion to reduce the amount of labeled data required to train the model without compromising performance~\cite{yoon_vae, ellefsen_ga}.

All the aforementioned studies contributed significantly to improving the state\nobreakdash-of\nobreakdash-the\nobreakdash-art of RUL estimation, however none of them properly addressed the problem of quantifying the uncertainty associated with the predictions. It is only in the last three years that Bayesian deep learning has started to draw the attention of the research community. For example, Kraus et al.~\cite{kraus_bayesian} developed a combined approach consisting of a non\nobreakdash-parametric explicit lifetime model, a Bayesian linear model, and a Bayesian recurrent neural network trained via Bayes by Backprop that achieved competitive performance on the C\nobreakdash-MAPSS dataset with the additional advantage of being interpretable and providing confidence intervals around predictions.
Another Bayesian method is that of Peng et al.~\cite{peng_bayesian}, who used Bayes by Backprop and Monte Carlo dropout to train a Bayesian multiscale convolutional neural network and a Bayesian bidirectional long short\nobreakdash-term memory network on the ball bearing dataset from the IEEE PHM 2012 data challenge~\cite{nectoux_phm12, bearing_nectoux} and the C\nobreakdash-MAPSS dataset, respectively. Experimental results showed superior performance compared to their frequentist counterparts trained via backpropagation.
Huang et al.~\cite{huang_bayesian} later proposed a Bayesian deep learning framework for capturing both the uncertainty in the model and the noise in the data. In particular, they assumed the RUL to follow a normal distribution parameterized by a Bayesian dense neural network trained via Bayes by Backprop.
A similar setup was adopted by Caceres et al.~\cite{caceres_bayesian}, who replaced the model by Huang et al. with a Bayesian recurrent neural network trained via both Monte Carlo dropout and Bayes by Backprop with flipout~\cite{wen_flipout}.
Li et al.~\cite{li_bayesian} further extended Huang et al.'s work by experimenting with different RUL distributions such as normal, logistic, and Weibull~\cite{lai_weibull}, parameterized by a Bayesian gated recurrent unit network~\cite{cho_gru} trained via Monte Carlo dropout. Furthermore, they presented a sequential Bayesian boosting procedure to enhance the predictive accuracy, which was validated on real\nobreakdash-world circuit breaker run\nobreakdash-to\nobreakdash-failure data.

Although not as popular as Bayes by Backprop and Monte Carlo dropout, Markov chain Monte Carlo was also used in predictive maintenance. Benker et al.~\cite{benker_bayesian} recently trained dense and convolutional Bayesian neural networks via Hamiltonian Monte Carlo~\cite{duane_hmc} and Bayes by Backprop on the C\nobreakdash-MAPSS dataset with some success. Moreover, they leveraged the uncertainty information obtained from the Bayesian models to further reduce the error rate. Despite improving over vanilla Markov chain Monte Carlo, such an algorithm still suffers from efficiency issues in big data settings due to the high computational cost of simulating Hamiltonian dynamics.
On the contrary, the method we utilize -- Stein variational gradient descent -- is both expressive like Markov chain Monte Carlo and efficient like backpropagation.

%------------------------------------------------------------------------
\section{Methodology}
\label{sec:methodology}

\subsection{Bayesian Neural Networks}
\label{subsec:bayesian_neural_networks}
In frequentist neural networks uncertainty is totally disregarded: weights are assumed to be deterministic and their optimal values are learned directly from data \text{$\mathcal{D} = \{(\mathbf{x}_i, y_i)\}_{i=1}^N$} via maximum likelihood (or maximum a posteriori when a regularization term is included) estimation.
On the contrary, in Bayesian neural networks weights are assumed to be random variables with a {prior distribution} $p(\mathbf{w})$ that explicitly encodes domain knowledge about the task at hand, and a {posterior distribution} $p(\mathbf{w}\mid\mathcal{D})$, i.e. the conditional distribution that results from updating $p(\mathbf{w})$ with information from the {likelihood} $p(\mathcal{D}\mid\mathbf{w})$~\cite{jospin_tutorial}. Following Bayes' rule, the posterior can be calculated as:
\begin{equation}
\label{eq:posterior}
p(\mathbf{w}\mid\mathcal{D}) = \frac{p(\mathcal{D}\mid\mathbf{w})p(\mathbf{w})}{p(\mathcal{D})},
\end{equation}
where $p(\mathcal{D}) = \int_{\mathbf{w'}} p(\mathcal{D}\mid\mathbf{w'})p(\mathbf{w'}) d\mathbf{w'}$ is known as the {evidence}. Knowing the posterior enables us to determine the {posterior predictive distribution} $p(\hat{y}\mid\hat{\mathbf{x}}, \mathcal{D})$, i.e. the distribution of output $\hat{y}$ as a function of input $\hat{\mathbf{x}}$, which is calculated as the weighted mean of the likelihoods of the infinite ensemble of neural networks induced by \text{$p(\mathbf{w}\mid\mathcal{D})$}~\cite{weight_uncertainty}:
\begin{align}
\label{eq:posterior_predictive}
p(\hat{y}\mid\hat{\mathbf{x}},\mathcal{D}) &=
\mathbb{E}_{p(\mathbf{w}\mid\mathcal{D})}[p(\hat{y}\mid\hat{\mathbf{x}},\mathbf{w})] \nonumber \\
&=\int_{\mathbf{w}} p(\hat{y}\mid\hat{\mathbf{x}},\mathbf{w}) p(\mathbf{w}\mid\mathcal{D}) d\mathbf{w}.
\end{align}
In practice, we can approximate it using a finite number of Monte Carlo samples from the posterior and calculate useful statistics such as mean, variance, skewness, etc. In particular, the mean or the mode of the posterior predictive distribution is typically used as a prediction. Moreover, we can quantify the uncertainty in the neural network's weights (commonly referred to as {epistemic}), which is captured by a fraction of the predictive variance and can be lowered by collecting more data~\cite{kendall_uncertainty}. A further advantage of adopting a Bayesian perspective is that, since a prior is imposed on $\mathbf{w}$, regularization is naturally provided and overfitting is less of a concern even in low data regimes.

Even though in theory $p(\mathbf{w}\mid\mathcal{D})$ can be derived via exact Bayesian inference using \cref{eq:posterior}, this is mostly impossible due to the high dimensionality of $\mathbf{w}$ and a functional form of the neural network not amenable to integration~\cite{weight_uncertainty}. A possible solution is to approximate the posterior through variational inference~\cite{wainwright_vi}, a technique that substitutes the potentially multimodal and/or heavy\nobreakdash-tailed true posterior with a simpler surrogate distribution.

\subsection{Bayes by Backprop}
\label{subsec:bayes_by_backprop}
{Bayes by Backprop}~\cite{graves_variational, weight_uncertainty} is an iterative gradient\nobreakdash-based variational inference approach that assumes the surrogate posterior to be a parametric distribution \text{$q(\mathbf{w}\mid{\bm{\theta}})$} from a tractable family (typically a multivariate normal with diagonal covariance matrix). The goal is to learn the optimal parameters ${\bm{\theta}}^\star$ that minimize the \text{Kullback-Leibler} divergence (KL) of $p(\mathbf{w}\mid\mathcal{D})$ from \text{$q(\mathbf{w}\mid{\bm{\theta}})$}:
\begin{align}
{\bm{\theta}}^\star &= \arg\min_{\bm{\theta}}
\text{KL}[q(\mathbf{w}\mid{\bm{\theta}})
\mid \mid
p(\mathbf{w}\mid\mathcal{D})] \nonumber \\
&= \arg\min_{\bm{\theta}}
\int_{\mathbf{w}}
q(\mathbf{w}\mid{\bm{\theta}})
\log
\frac{q(\mathbf{w}\mid{\bm{\theta}})}
{p(\mathbf{w}) p(\mathcal{D}\mid\mathbf{w})}
\textrm{d}\mathbf{w} \nonumber \\
&= \arg\min_{\bm{\theta}}
\text{KL}
\left[
q(\mathbf{w}\mid{\bm{\theta}})
\mid \mid
p(\mathbf{w})
\right]
-
\mathbb{E}_{q(\mathbf{w}\mid{\bm{\theta}})}
\left[
\log
p(\mathcal{D}\mid\mathbf{w})
\right].
\end{align}
The resulting loss function is known as the {evidence lower bound} (ELBO) or the {variational free energy} (VFE) and explicitates the trade\nobreakdash-off between matching the simplicity of the prior (complexity loss) and the complexity of the data (likelihood loss)~\cite{neal_elbo}. It is denoted as
\begin{align}
\label{eq:elbo}
\mathcal{F}(\mathcal{D}, {\bm{\theta}}) =
\underbrace{\text{KL}
\left[	
q(\mathbf{w}\mid{\bm{\theta}})
\mid \mid
p(\mathbf{w})
\right]
}_\text{complexity loss}
-
\underbrace{
\mathbb{E}_{q(\mathbf{w}\mid{\bm{\theta}})}
\left[
\log
p(\mathcal{D}\mid\mathbf{w})
\right]
}_\text{likelihood loss}
.
\end{align}
Exactly minimizing the ELBO is computationally infeasible in many cases~\cite{weight_uncertainty}. Bayes by Backprop approximates it via Monte Carlo sampling:
\begin{align}
\label{eq:approx_elbo}
\mathcal{F}(\mathcal{D}, {\bm{\theta}}) \approx
\sum_{i=1}^M
\log q(\mathbf{w}_{i}\mid{\bm{\theta}})
-
\log p(\mathbf{w}_{i})
-
\log
p(\mathcal{D}\mid\mathbf{w}_{i}),
\end{align}
where $M$ denotes the number of Monte Carlo samples and $\mathbf{w}_{i}$ the \text{$i$-th} Monte Carlo sample drawn from \text{$q(\mathbf{w}\mid{\bm{\theta}})$} using the {reparameterization trick}~\cite{kingma_vae}. Gradient\nobreakdash-based optimization can be used to minimize \cref{eq:approx_elbo} and learn the optimal parameters of the surrogate posterior.

\subsection{Stein Variational Gradient Descent}
\label{subsec:stein_variational_gradient_descent}
{Stein variational gradient descent}~\cite{liu_svgd} is an iterative gradient\nobreakdash-based variational inference algorithm that assumes the surrogate posterior $q(\mathbf{w}\mid{\{\mathbf{w}_i\}_{i=1}^M})$ to be a non\nobreakdash-parametric distribution represented by a finite set of $M$ randomly initialized particles $\{\mathbf{w}_i\}_{i=1}^M$ that are gradually evolved toward the true posterior $p(\mathbf{w}\mid{\mathcal{D}})$ through a sequence of transforms. It has a simple form that is similar to standard gradient ascent, and equivalent to it when $M = 1$ (in this case it yields the maximum a posteriori estimate). As a result, it is very adaptable and scalable and readily usable in combination with other optimization techniques such as stochastic gradient ascent, momentum, and adaptive learning rates (e.g. Adam~\cite{adam}).
Let $\bm{\tau}: \mathbb{R}^{D} \rightarrow \mathbb{R}^{D}$ denote the transform applied to the particles at the $l$\nobreakdash-th iteration. The goal is to choose $\bm{\tau}$ such that it maximally reduces the Kullback\nobreakdash-Leibler divergence between the true posterior $p(\mathbf{w}\mid{\mathcal{D}})$ and the transformed surrogate posterior $q_{\bm{\tau}}(\mathbf{w}\mid{\{\mathbf{w}_i\}_{i=1}^M})$:
\begin{equation}
    \bm{\tau}^\star = \arg\min_{\bm{\tau}}
    \text{KL}[q_{\bm{\tau}}(\mathbf{w}\mid{\{\mathbf{w}_i\}_{i=1}^M})
    \mid \mid
    p(\mathbf{w}\mid{\mathcal{D}})].
\end{equation}
In order to do so, it is necessary to impose a few constraints on its functional form. First, we require $\bm{\tau}$ to be invertible, such that the density of the transformed variable can be easily computed via the change of variables formula. To ensure that, we define it as a small perturbation of the identity map:
\begin{equation}
\bm{\tau}(\mathbf{w}) = \mathbf{w} + \epsilon \bm{\phi}(\mathbf{w}),
\end{equation}
where $\epsilon \in \mathbb{R}$ denotes the perturbation magnitude and $\bm{\phi}: \mathbb{R}^{D} \rightarrow \mathbb{R}^{D}$ the perturbation direction. If $\lvert \, \epsilon \, \rvert$ is sufficiently small and $\bm{\phi}$ is smooth, then $\bm{\tau}$ is invertible.
Second, we restrict $\bm{\phi}$ to be within the unit ball of a $D$\nobreakdash-dimensional reproducing kernel Hilbert space~\cite{berlinet_rkhs} induced by a positive\nobreakdash-definite kernel \text{$k: \mathbb{R}^{D} \times \mathbb{R}^{D} \rightarrow \mathbb{R}$}.
%(which is a hyperparameter of the algorithm together with the number of particles).
Given these conditions, the following closed\nobreakdash-form expression holds for the direction of steepest descent of \text{$\text{KL}[q_{\bm{\tau}}(\mathbf{w}\mid{\{\mathbf{w}_i\}_{i=1}^M}) \mid \mid p(\mathbf{w}\mid{\mathcal{D}})]$} at the $l$\nobreakdash-th iteration:
\begin{align}
\label{eq:svgd}
\bm{{\phi}^\star}(\mathbf{w}^{(l)}_i) &= \frac{1}{M}
\sum_{j=1}^M \underbrace{k(\mathbf{w}^{(l)}_j, \mathbf{w}^{(l)}_i) \nabla_{\mathbf{w}^{(l)}_j} \log \hat{p}(\mathbf{w}^{(l)}_j\mid\mathcal{D})}_\text{driving force} \nonumber \\
&+ \frac{1}{M} \sum_{j=1}^M \underbrace{\nabla_{\mathbf{w}^{(l)}_j} k(\mathbf{w}^{(l)}_j, \mathbf{w}^{(l)}_i)}_\text{repulsive force},
\end{align}
where $\hat{p}(\mathbf{w}\mid\mathcal{D}) = p({\mathcal{D}}\mid\mathbf{w}) p(\mathbf{w})$ denotes the unnormalized weight posterior. The first term represents a driving force that guides the particles toward high probability regions of $p(\mathbf{w}\mid{\mathcal{D}})$, and the second a repulsive force that prevents the particles from collapsing into local modes of $p(\mathbf{w}\mid{\mathcal{D}})$. The updated weights are then calculated as
\begin{equation}
    \mathbf{w}^{(l + 1)}_i = \mathbf{w}^{(l)}_i + \epsilon^{(l)} \bm{\hat{\phi}^\star}(\mathbf{w}^{(l)}_i),
\end{equation}
where $\epsilon^{(l)}$ denotes the perturbation magnitude at the $l$\nobreakdash-th iteration. It is straightforward to derive a corresponding loss function from \cref{eq:svgd} and apply standard gradient\nobreakdash-based optimization to learn the optimal particle\nobreakdash-based approximation of the true posterior.

\subsection{Uncertainty-Informed Predictions}
\label{subsec:uncertainty_informed_predictions}
The Bayesian theoretical framework offers a natural way to reason about uncertainty in predictions, which is crucial for informed decision-making in real-world applications. For example, if the RUL estimate provided by the model is highly uncertain (i.e. the predictive variance is large), a machine operator might opt to preemptively schedule maintenance before reaching the predicted threshold, since the costs associated with unforeseen equipment breakdowns generally exceed the expenses incurred through underutilization. We incorporate this conservative strategy directly into the prognostic procedure by proposing a simple heuristic that helps preventing late predictions. Given a test sample $\hat{\mathbf{x}}$, let $\mu(\hat{{y}})$ and $\sigma(\hat{{y}})$ denote the mean and the standard deviation of the posterior predictive distribution $p(\hat{y}\mid\hat{\mathbf{x}},\mathcal{D})$, respectively. Assuming that $\mu(\hat{{y}})$ represents the RUL estimate (as mentioned in \cref{subsec:bayesian_neural_networks}, the mode of the posterior predictive distribution could be an alternative), we propose the following modification:
\begin{equation}
\label{eq:heuristic}
    \mu^{\ast}(\hat{{y}}) = \mu(\hat{{y}}) - p_{\text{late}} \, k \, \sigma(\hat{{y}}),
\end{equation}
where $\mu^{\ast}(\hat{{y}})$ is the corrected RUL estimate, $p_{\text{late}} \in [0, 1]$ the probability of late prediction, and $k > 0$ a hyperparameter that controls the correction strength. If $p_{\text{late}} = 0$, than the model is risk-averse and no correction is necessary. Conversely, if $p_{\text{late}} = 1$, the model is risk-seeking, hence a correction proportional to $k$ is applied. The probability of late prediction $p_{\text{late}}$ can be estimated on a subset of held-out data {$\tilde{\mathcal{D}} = \{(\tilde{\mathbf{x}}_i, \tilde{y}_i)\}_{i=1}^{\tilde{N}}$} as
\begin{equation}
\label{eq:p_late}
    p_{\text{late}} = \frac{1}{\tilde{N}} \sum_{i=1}^{\tilde{N}} \mathds{1}(\mu(\hat{\tilde{{y}}}_i) > \tilde{y}_i),
\end{equation}
where $\mathds{1}(\cdot)$ denotes the indicator function. It is worth noting that our method can be easily adjusted as new data are acquired by reestimating $p_{\text{late}}$ on such data. This adaptability could also prove beneficial in addressing scenarios characterized by domain shift.

%------------------------------------------------------------------------
\section{Experimental Setup}
\label{sec:experimental_setup}

\subsection{Dataset}
\label{subsec:dataset}
The proposed Bayesian methods are evaluated using the simulated turbofan engine degradation data from the publicly accessible NASA's C\nobreakdash-MAPSS dataset\footnote{\href{https://www.nasa.gov/content/prognostics-center-of-excellence-data-set-repository}{https://www.nasa.gov/content/prognostics-center-of-excellence-data-set-repository}}~\cite{saxena_cmapss, damage_propagation}, which comprises $4$ subsets of multivariate trajectories from $21$ sensors. Every subset includes a training set and a test set. The training set consists of run\nobreakdash-to\nobreakdash-failure sensor recordings of several engines acquired under various operating conditions and fault modes.
The engines are assumed to be healthy at first, however their initial wear and manufacturing variations are unknown. They deteriorate over time until a system failure occurs, with the last data point corresponding to the time cycle when the unit fails. On the contrary, sensor records in the test set end some time prior to the failure. The objective is to estimate the RUL of each engine in the test set. The true RUL values of the test engines are provided for performance evaluation. Each subset has $26$ columns: engine number, time cycle, $3$ operational settings (which define the operating condition), and $21$ sensor measurements. Detailed information about the $4$ subsets, identified as FD001, FD002, FD003, and FD004 is shown in \cref{tab:cmapss}.

\subsection{Preprocessing}
\label{subsec:preprocessing}

\paragraph{Feature selection}
\label{par:feature_selection}
In FD001 and FD003, sensors $1$, $5$, $6$, $10$, $16$, $18$, and $19$ are constant throughout the engine's lifespan, hence they do not provide any meaningful information for predicting the RUL. Furthermore, FD001 and FD003 are subjected to a single operating condition. As a result, in FD001 and FD003 only $14$ of the $21$ sensors are utilized as input features, with indices $2$, $3$, $4$, $7$, $8$, $9$, $11$, $12$, $13$, $14$, $15$, $17$, $20$ and $21$, while the $3$ operational settings are ignored. In contrast, there are no constant measurements in FD002 and FD004, and the existence of $6$ operating conditions makes it more difficult to detect deterioration patterns. Therefore, in FD002 and FD004 all the $21$ sensors and the $3$ operational settings are employed as input features.

\paragraph{Normalization}
\label{par:normalization}
The collected sensor data and operational settings are normalized to be within the interval $\left[-1, 1\right]$ by means of {min\nobreakdash-max} normalization:
\begin{equation}
x_{ij}^{norm} = \frac{2 (x_{ij} - x_{j}^{min})}{x_{j}^{max} - x_{j}^{min}} - 1, \ \ \forall \ i, j,
\end{equation}
where $x_{ij}$ denotes feature $j$ at time step $i$, $x_{ij}^{norm}$ the normalized value of $x_{ij}$, $x_{j}^{max}$ the maximum value of feature $j$ across time steps of all trajectories and $x_{j}^{min}$ the minimum value.

\begin{figure}[t]
    \begin{center}
        \includegraphics[height=.40\textheight]{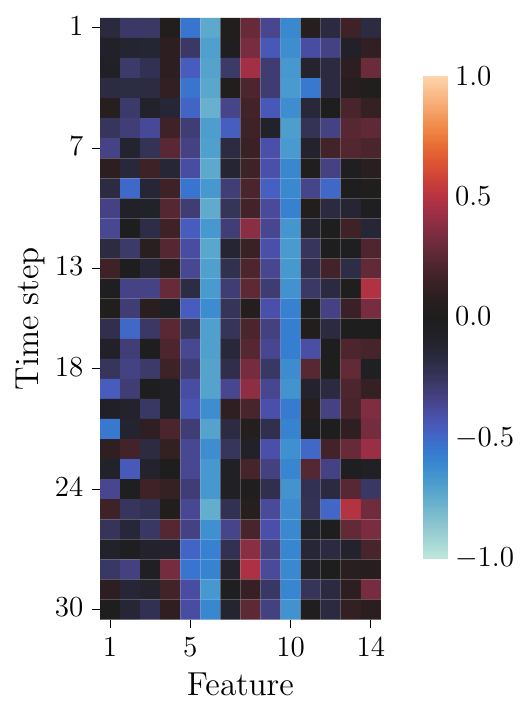}
        \caption{Min\nobreakdash-max normalized sample from FD001 test set with $T = 30$ and $F = 14$.
        }
        \label{fig:data_fd001_test_27}
    \end{center}
\end{figure}

\paragraph{Sliding window segmentation}
\label{par:sliding_window_segmentation}
We can generally extract more information from a multivariate trajectory by examining the temporal sequence data as opposed to only considering the individual data points at each time step. To retain the deterioration patterns hidden in the time dimension, the sliding window segmentation method originally proposed by Babu et al.~\cite{rul_cnn_first} is used to partition the trajectory into overlapping segments of fixed size. Let $T$ denote the window size and $F$ the number of features.
At each time step, all future sensor records within the window are gathered into a $T \times F$ matrix (see \cref{fig:data_fd001_test_27}) such that from a trajectory of length $L$ exactly $L - T + 1$ segments are extracted. If $L < T$, the trajectory is discarded. The resulting segments are then labeled with the RUL of the last data point in the window and given as input to the deep learning models described in \cref{subsec:deep_learning_models}.

\paragraph{Target rectification}
\label{par:target_rectification}
A popular technique in the literature~\cite{gugulothu_rnn, zheng_lstm, rul_cnn_first, rul_dcnn, benker_bayesian} is to use a piece\nobreakdash-wise linear degradation model in which the RUL target function is assumed to be constant until a threshold value, $R_{early}$, beyond which it linearly decreases to $0$. From an implementation point of view, this means rectifying the targets, i.e. setting the RUL to $R_{early}$ for all samples whose RUL is larger than $R_{early}$. The intuition behind it is that a system generally operates correctly at first and starts to degrade only after a certain amount of wear. Following the aforementioned literature, we set $R_{early} = 125$ for both the training and the test set.
Hyperparameters of the preprocessing phase are reported in \cref{tab:preprocessing}.

\input{dataset.tex}

\input{preprocessing.tex}

\subsection{Deep Learning Models}
\label{subsec:deep_learning_models}

\input{hyperparameters.tex}

The goal of this work is to explore the use of Stein variational gradient descent to train Bayesian deep learning models for RUL estimation. More specifically, we investigate whether it converges faster and yields better predictive accuracy than a competitive baseline such as Bayes by Backprop. To this end, we implement the two simple yet effective neural network architectures proposed by Benker et al.~\cite{benker_bayesian}, train them under identical experimental conditions with both methods, and compare their performance. For the sake of completeness, we also consider backpropagation as a simpler baseline for training the frequentist counterparts of the selected models, which are defined as follows:

\begin{itemize}
    \item Dense3 (D3): dense neural network architecture originally proposed by Benker et al. The $T \times F$ input matrix, with $T$ denoting the window size and $F$ the number of features, is flattened into a $1$\nobreakdash-dimensional vector, which is then fed to $3$ consecutive $100$ neurons dense layers. A final output neuron returns the RUL prediction. After each dense hidden layer, a sigmoid activation function is applied.
    \item Conv2Pool2 (C2P2): convolutional neural network architecture originally proposed by Babu et al.~\cite{rul_cnn_first} and later used by Benker et al. too. The $T \times F$ input matrix is fed to a $5 \times 14$ convolutional layer with $8$ channels followed by a $2 \times 1$ average pooling layer. A second $2 \times 1$ convolution with $14$ channels is applied, followed by another $2 \times 1$ average pooling. The output is flattened into a $1$\nobreakdash-dimensional vector and a final output neuron returns the RUL prediction. After each convolutional hidden layer, a sigmoid activation function is applied.
\end{itemize}

All models are trained for $50$ epochs using as a negative log\nobreakdash-likelihood Huber loss~\cite{huber_loss} with a residual threshold $\delta$ arbitrarily set to $100$ (which is the overall cost to minimize for frequentist models and only a fraction of it for Bayesian ones, see \cref{eq:approx_elbo} and \cref{eq:svgd}), which is less sensitive to outliers than plain mean squared error. For each epoch, training samples and their corresponding targets are randomly split into multiple batches of size $512$ and fed to the model. The neural network's weights are updated based on the gradient of the loss function computed batchwise by means of Adam~\cite{adam} optimizer, with a learning rate of $0.01$, reduced by a factor of $10$ after $40$ epochs for fine\nobreakdash-tuning.
For frequentist variants trained via backpropagation, weights are initialized according to Kaiming's uniform scheme~\cite{kaiming_init} and dropout with a drop probability of $0.2$ is applied to avoid overfitting.
For Bayesian variants trained via Bayes by Backprop, prior and surrogate posterior are modeled as zero\nobreakdash-mean multivariate normal distributions with diagonal covariance matrices and per\nobreakdash-dimension initial standard deviations equal to $0.1$ and $\mathrm{softplus}(\rho) = \ln (1 + e^\rho)$ with $\rho = 1$, respectively. Note the use of softplus\nobreakdash-reparameterization as per reference~\cite{weight_uncertainty} to ensure a strictly positive standard deviation during training.
The same prior is utilized for Bayesian variants trained via Stein variational gradient descent. In this case, the surrogate posterior is approximated by $10$ particles, initialized by drawing that number of samples from the prior. As a kernel, we employ a radial basis function with bandwidth chosen using the median\nobreakdash-based heuristic by Liu et al.~\cite{liu_svgd}.
As mentioned in \cref{subsec:uncertainty_informed_predictions}, we use the mean of the posterior predictive distribution as the RUL estimate. For Bayesian models, we correct the prediction according to \cref{eq:heuristic}, where we set the correction factor $k = 1$. The probability of late prediction $p_{\text{late}}$ is computed on the training set (since the C\nobreakdash-MAPSS dataset does not include a validation set) based on \cref{eq:p_late}.
Hyperparameters of the training algorithms are summarized in \cref{tab:hyperparameters}.

\subsection{Performance Metrics}
\label{subsec:performance_metrics}
\begin{figure}[t]
    \begin{center}
        \includegraphics[height=.40\textheight]{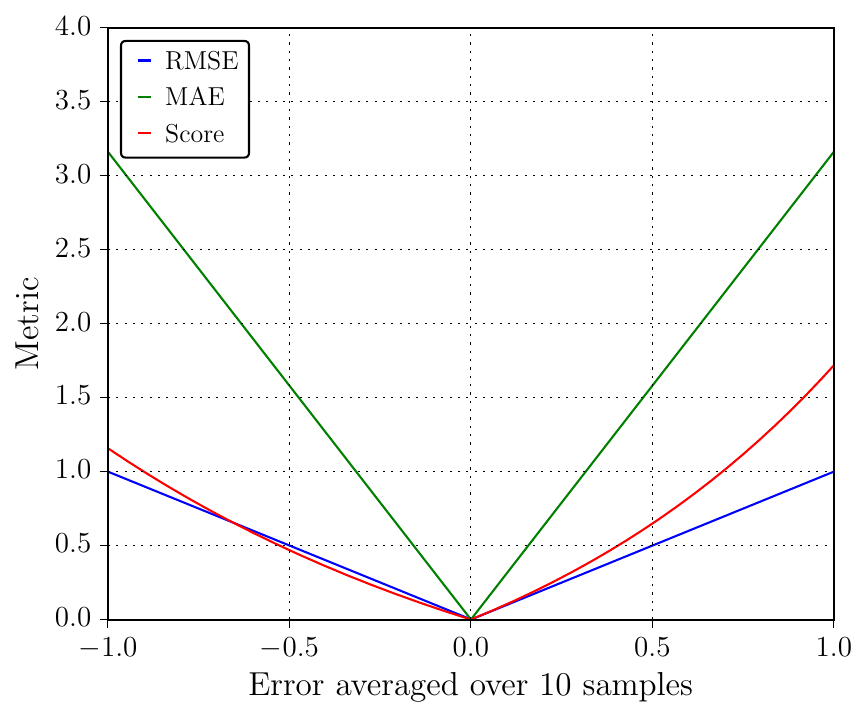}
        \caption{Comparison between $\text{RMSE}$, $\text{MAE}$, and score function with respect to different average error values, where the error is the difference between the prediction and the target. The score function is asymmetric: it penalizes late predictions ($\text{error} > 0$) more than early ones ($\text{error} < 0$).
        }
        \label{fig:metrics_comparison}
    \end{center}
\end{figure}

For performance evaluation, we calculate root mean squared error (RMSE), mean absolute error (MAE) and the score function (Score) proposed by Saxena et al.~\cite{damage_propagation}:
\begin{equation}
\text{RMSE} = \sqrt{\frac{1}{N} \sum_{i=1}^{N}{d_i^2}},
\end{equation}
\begin{equation}
\text{MAE} = \frac{1}{N} \sum_{i=1}^{N}{\lvert d_i \rvert},
\end{equation}
\begin{equation}
\text{Score} = 
\sum_{i=1}^{N}{e^{s_i} - 1}, \ \ \text{where} \ \ s_i = 
\begin{cases}
-\frac{d_i}{13}, & \text{for} \ \ d_i < 0 \\
\hphantom{-} \frac{d_i}{10}, & \text{for} \ \ d_i \geq 0 \\
\end{cases},
\end{equation}
where $N$ is the number of test samples and $d_i$ the error, i.e. the difference between estimated and true RUL values for the $i$\nobreakdash-th test sample. Good models should achieve relatively low values on all those metrics. 
As shown in \cref{fig:metrics_comparison}, $\text{RMSE}$ and $\text{MAE}$ equally penalize early and late predictions, while the asymmetric score function penalizes late predictions more than early ones. Minimizing the score is crucial as late predictions often result in more catastrophic consequences than early ones because maintenance will be planned too late. Nonetheless, it is useful to monitor $\text{RMSE}$ and $\text{MAE}$ too, since the score is particularly sensitive to outliers due to the exponentiation.

\subsection{Implementation and Hardware}
\label{subsec:implementation_and_hardware}
Software for the experimental evaluation was implemented in Python 3.8.13. In particular, we used NumPy 1.23.4\footnote{\href{https://github.com/numpy/numpy/tree/v1.23.4}{https://github.com/numpy/numpy/tree/v1.23.4}}~\cite{harris_numpy} to preprocess the data, PyTorch 1.12.1\footnote{\href{https://github.com/pytorch/pytorch/tree/v1.12.1}{https://github.com/pytorch/pytorch/tree/v1.12.1}}~\cite{pytorch_paszke} to implement the deep learning models and the training loops, BayesTorch 0.0.1\footnote{\href{https://github.com/lucadellalib/bayestorch/tree/v0.0.1}{https://github.com/lucadellalib/bayestorch/tree/v0.0.1}} to implement the Bayesian inference algorithms, Ray AIR 2.0.1\footnote{\href{https://github.com/ray-project/ray/tree/ray-2.0.1}{https://github.com/ray-project/ray/tree/ray-2.0.1}}~\cite{ray_moritz, tune_liaw} for experiment execution, and Matplotlib 3.6.2\footnote{\href{https://github.com/matplotlib/matplotlib/tree/v3.6.2}{https://github.com/matplotlib/matplotlib/tree/v3.6.2}}~\cite{hunter_matplotlib} and Seaborn 0.12.1\footnote{\href{https://github.com/mwaskom/seaborn/tree/v0.12.1}{https://github.com/mwaskom/seaborn/tree/v0.12.1}}~\cite{waskom_seaborn} for plotting.
All the experiments were run on an Ubuntu 20.04.5 LTS machine with an Intel i7\nobreakdash-10875H CPU with $8$ cores @ $2.30$ GHz, $32$ GB RAM and an NVIDIA GeForce RTX $3070$ GPU @ $8$ GB with CUDA Toolkit 11.3.1.

\section{Experimental Results and Discussion}
\label{sec:experimental_results_and_discussion}
\begin{figure*}[hbt!]
    \centering
    \begin{subfigure}{0.81\textwidth}
      \includegraphics[width=0.49\textwidth]{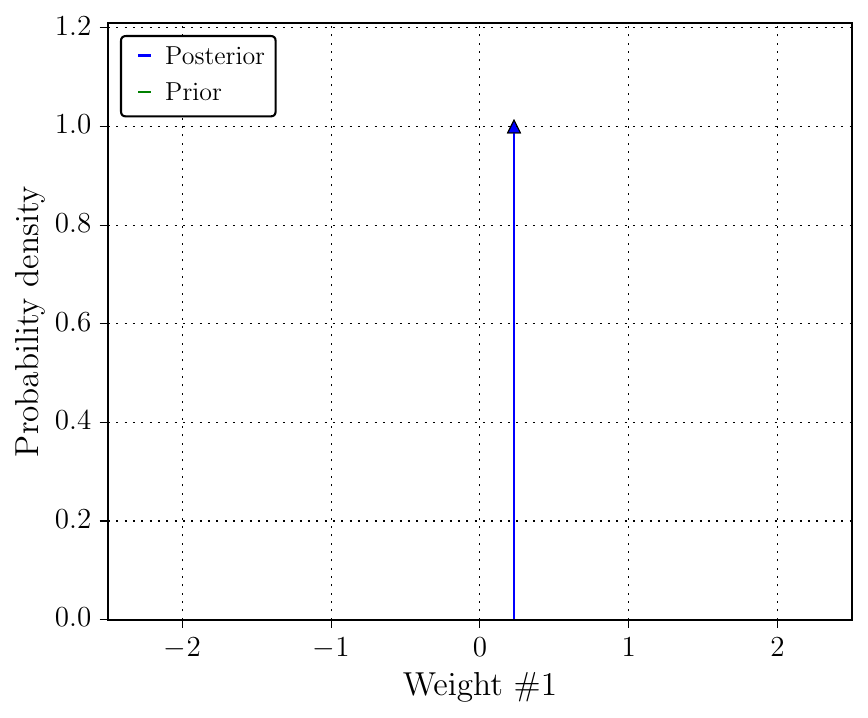}
      \includegraphics[width=0.49\textwidth]{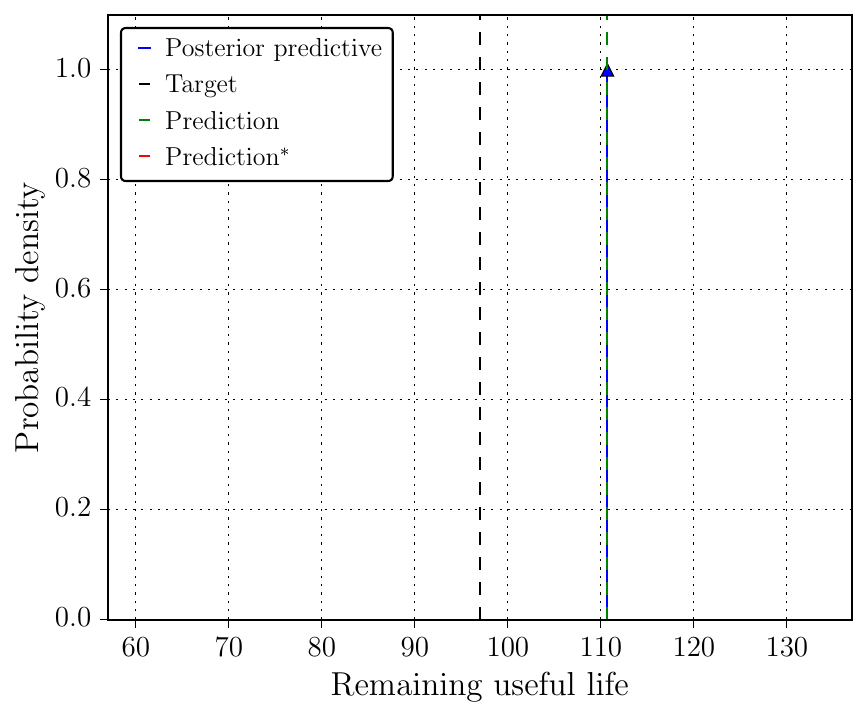}
      \vspace{-2mm}
      \caption{BP-D3}
    \end{subfigure}
    \begin{subfigure}{0.81\textwidth}
      \vspace{2mm}
      \includegraphics[width=0.49\textwidth]{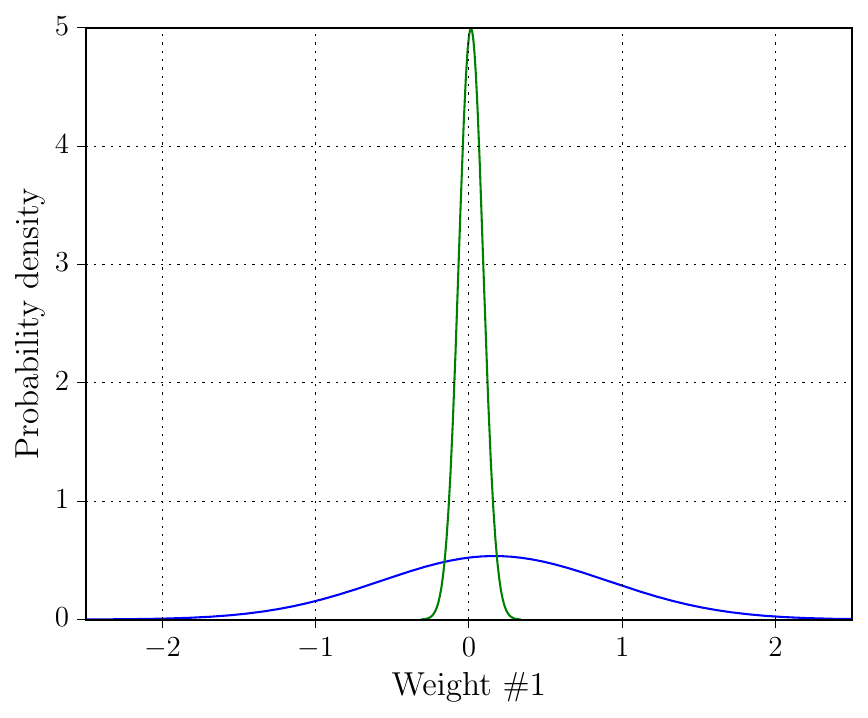}
      \includegraphics[width=0.49\textwidth]{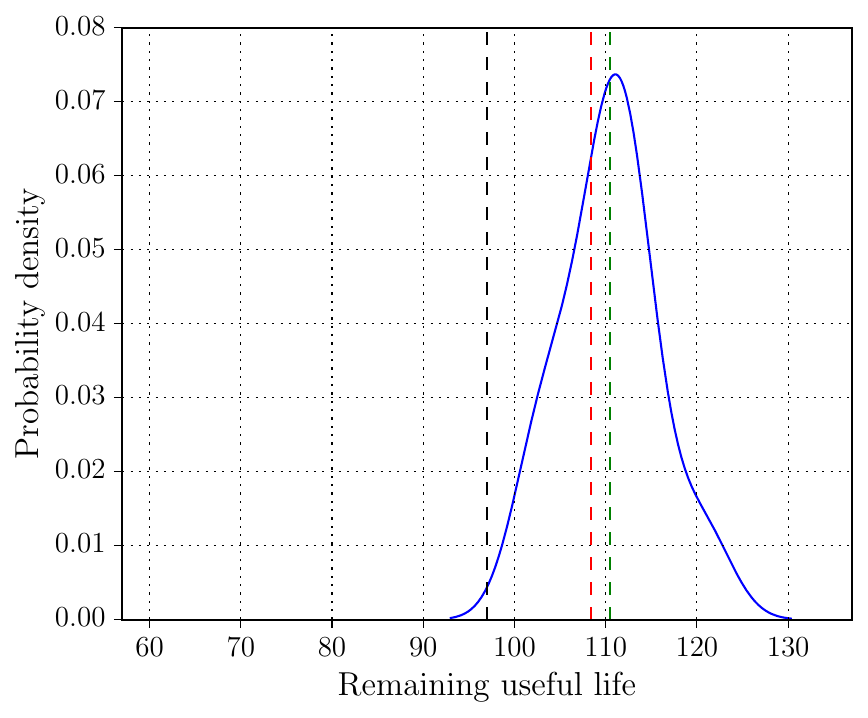}
      \vspace{-2mm}
      \caption{BBB-D3}
    \end{subfigure}
    \begin{subfigure}{0.81\textwidth}
      \vspace{2mm}
      \includegraphics[width=0.49\textwidth]{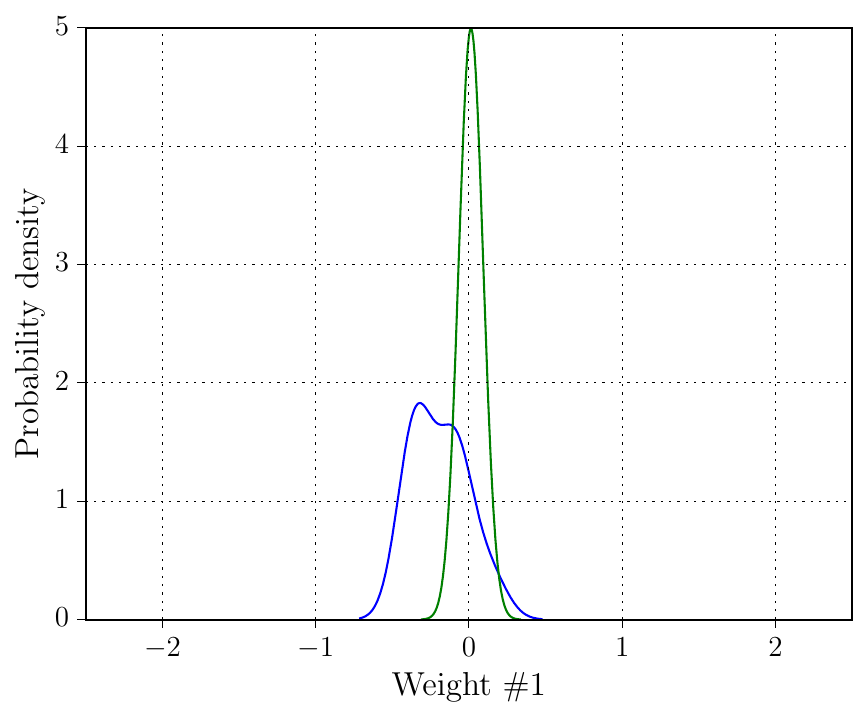}
      \includegraphics[width=0.49\textwidth]{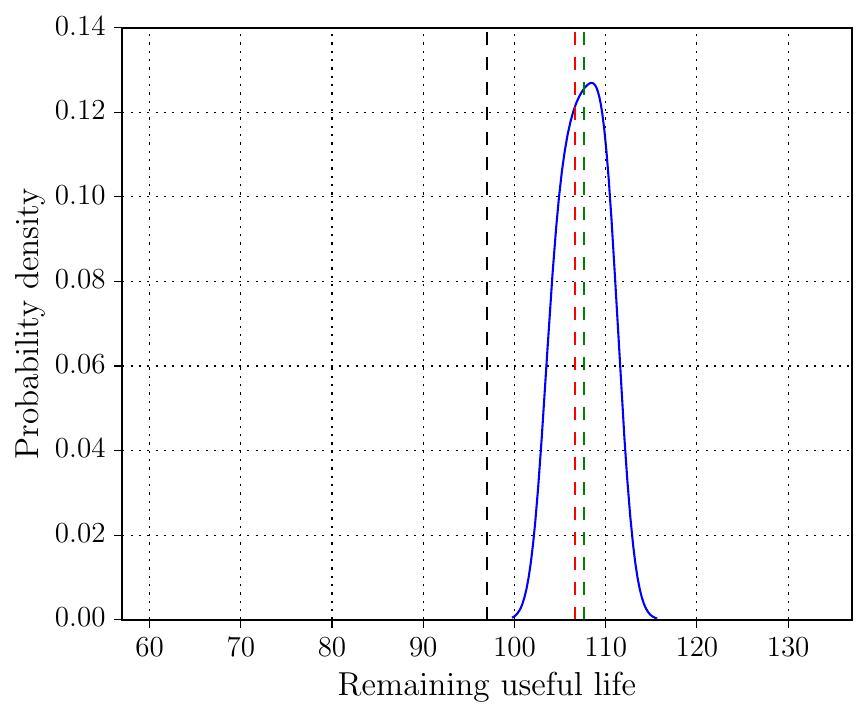}
      \vspace{-2mm}
      \caption{SVGD-D3}
    \end{subfigure}
\caption{
Prior and posterior distributions (first column) of the first weight of the first hidden layer and corresponding posterior predictive distribution (second column) of a test sample of Dense3 (D3) trained on FD001 via backpropagation (BP), Bayes by Backprop (BBB), and Stein variational gradient descent (SVGD). Prediction\textsuperscript{$\ast$} is computed using the heuristic described in \cref{subsec:uncertainty_informed_predictions}.
}
\label{fig:distributions}
\end{figure*}

Experimental results for D3 and C2P2 trained via backpropagation (BP), Bayes by Backprop (BBB) and Stein variational gradient descent (SVGD) reported in \cref{tab:performance} show that Stein variational gradient descent yields lower error rates than the other two approaches, with D3\nobreakdash-SVGD performing the best across all subsets.
In particular, as illustrated in \cref{fig:distributions}, Stein variational gradient descent tends to produce a narrower posterior and posterior predictive distribution with mean closer to the true RUL value compared to Bayes by Backprop. The lower variance in the predictions is a direct consequence of the fact that the neural network is more certain about the values of its weights, suggesting that Stein variational gradient descent converged faster and to a more accurate solution than Bayes by Backprop.
This is confirmed by the generally lower standard deviation values in the performance metrics of D3\nobreakdash-SVGD and C2P2\nobreakdash-SVGD, which is a further indication of a less brittle optimization process.

\input{results.tex}

Despite the differences between Bayes by Backprop and Stein variational gradient descent, we observe that, across all the subsets, Bayesian deep learning models trained with these methods perform better than their frequentist counterparts and especially D3\nobreakdash-BP, which fails to achieve good performance in FD002 and FD004. This stabilizing effect is likely due to the implicit ensemble learning property that characterizes Bayesian neural networks. However, contrarily to Benker et al.~\cite{benker_bayesian}, we did not find Bayes by Backprop to clearly improve over backpropagation with respect to the score, which might be imputable to a different hyperparameter configuration.

Regarding the comparison between different subsets, we notice that the predictive performance varies drastically, with a substantial drop in FD002 and FD004 for all the configurations. This is in line with expectations since detecting deterioration patterns is more difficult in those subsets due to the presence of $6$ instead of $1$ operating conditions. A performance gap exists between D3 and C2P2 too. In fact, the latter performs slightly worse than D3 on the more complex FD002 and FD004, and significantly worse on the less challenging FD001 and FD003. This might be due to overfitting or to a suboptimal neural network architecture.
The evaluation of more competitive deep learning models is beyond the scope of this research and will be addressed in future work.

Lastly, we emphasize that our uncertainty\nobreakdash-based heuristic consistently improves the score metric across all models. For instance, in the case of D3\nobreakdash-SVGD, the score on FD001 improves from $334$ to $318$. While a similar improvement is frequently observed for RMSE and MAE, it is less consistent, reflecting the fact that our method is primarily designed to mitigate the risk associated with late predictions rather than targeting error reduction.

%------------------------------------------------------------------------
\section{Conclusions and Future Work}
\label{sec:conclusions_and_future_work}
In this work Stein variational gradient descent was successfully applied to the task of RUL estimation for the first time. Bayesian dense and convolutional neural networks were trained using such method and compared to both the same models trained via Bayes by Backprop and their frequentist counterparts trained via backpropagation. Experiments on the simulated run\nobreakdash-to\nobreakdash-failure turbofan engine degradation data of the C\nobreakdash-MAPSS dataset showed that Bayesian deep learning models trained via Stein variational gradient descent consistently outperformed the other two approaches. Furthermore, a heuristic to reduce the risk of late predictions and enhance the performance based on the uncertainty information provided by the Bayesian models was proposed.
However, further investigations are necessary to harness the full potential of this technique.
First, the impact of the number of particles and the choice of the kernel should be analyzed.
Second, state\nobreakdash-of\nobreakdash-the\nobreakdash-art transformer\nobreakdash-based architectures should be included in the comparative study.
Lastly, extensions to Stein variational gradient descent such as neural variational gradient descent~\cite{langosco_nvgd}, which eliminates the need for a kernel, could be explored.

\section*{CRediT Author Statement}
% see https://www.elsevier.com/authors/policies-and-guidelines/credit-author-statement
\noindent \textbf{Luca Della Libera}: Conceptualization, Investigation, Methodology, Software, Validation, Visualization, Writing - Original Draft.\\
\textbf{Jacopo Andreoli}: Investigation, Methodology, Software, Validation, Visualization, Writing - Original Draft.\\
\textbf{Davide Dalle Pezze}: Methodology, Validation, Writing - Review \& Editing.\\
\textbf{Mirco Ravanelli}: Funding acquisition, Resources, Supervision, Writing - Review \& Editing.\\
\textbf{Gian Antonio Susto}: Funding acquisition, Resources, Supervision, Writing - Review \& Editing.

\section*{Declaration of Competing Interest}
The authors declare that there are no conflicts of interest.

\section*{Acknowledgments}
We thank Francesco Paissan for valuable mathematical discussions, and Domenico Lopez and Amrit Singh for software testing.
This work was supported by the Natural Sciences and Engineering Research Council of Canada (NSERC) and the Digital Research Alliance of Canada (alliancecan.ca). For GA Susto, this study was partially carried out within the MICS (Made in Italy – Circular and Sustainable) Extended Partnership and received funding from Next-GenerationEU (Italian PNRR – M4 C2, Invest 1.3 – D.D. 1551.11-10-2022, PE00000004). Moreover, this study was also partially carried out within the PNRR research activities of the consortium iNEST (Interconnected Nord-Est Innovation Ecosystem) funded by Next-GenerationEU (Italian PNRR – M4 C2, Invest 1.5 – D.D. 1058.23-06-2022, ECS00000043).

\bibliographystyle{elsarticle-num} 
\bibliography{bibliography}





% \end{thebibliography}
\end{document}

%% file: dataset.tex
\begin{table}[t]
	\setlength{\tabcolsep}{0.40pc}
	\centering 
	\begin{threeparttable}
		\caption{C-MAPSS dataset.}
		\label{tab:cmapss}
		\begin{tabular}{l@{\hskip 3pt}rrrr}
			\toprule
			\multirow{2.25}{*}{} &
			\multicolumn{4}{c}{C-MAPSS} \\
			\cmidrule(r){2-5} &
			{FD001} & {FD002} & {FD003} & {FD004} \\
			\midrule
			Training trajectories
			\multirow{1}{*} & $100$ & $260$ & $100$ & $249$ \\
			Test trajectories
			\multirow{1}{*} & $100$ & $259$ & $100$ & $248$ \\
            Operating conditions
            \multirow{1}{*} & $1$ & $6$ & $1$ & $6$ \\
            Fault modes
            \multirow{1}{*} & $1$ & $1$ & $2$ & $2$ \\
			\bottomrule
		\end{tabular}
	\end{threeparttable}
\end{table}

%% file: preprocessing.tex
\begin{table}[t]
	\setlength{\tabcolsep}{0.40pc}
	\centering 
	\begin{threeparttable}
		\caption{Hyperparameters of the preprocessing phase.}
		\label{tab:preprocessing}
		\begin{tabular}{l@{\hskip 3pt}rrrr}
			\toprule
			\multirow{2.25}{*}{} &
			\multicolumn{4}{c}{C-MAPSS} \\
			\cmidrule(r){2-5} &
			{FD001} & {FD002} & {FD003} & {FD004} \\
			\midrule
			$T$
			\multirow{1}{*} & $30$ & $20$ & $30$ & $15$ \\
			$F$
			\multirow{1}{*} & $14$ & $24$ & $14$ & $24$ \\
            Training samples
            \multirow{1}{*} & $\numprint{17731}$ & $\numprint{48819}$ & $\numprint{21820}$ & $\numprint{57763}$ \\
            Test samples
            \multirow{1}{*} & $100$ & $259$ & $100$ & $248$ \\
            $R_{early}$
            \multirow{1}{*} & $125$ & $125$ & $125$ & $125$ \\
			\bottomrule
		\end{tabular}
	\end{threeparttable}
\end{table}

%% file: hyperparameters.tex
\begin{table}[hbt!]
	\setlength{\tabcolsep}{0.40pc}
	\centering
	\begin{threeparttable}
		\caption{Hyperparameters of the training algorithms. $\bm{0}$ denotes the origin in $\mathbb{R}^D$ and $\mathds{I}$ the identity matrix in $\mathbb{R}^D$, where $D$ is the number of weights in the neural network.}
		\label{tab:hyperparameters}
		\begin{tabular}{l@{\hskip 3pt}r}
			\toprule
			%\multicolumn{1}{c}{} & \multicolumn{1}{c}{C-MAPSS} \\
			%\cmidrule{2-2}
			\multicolumn{1}{l}{Hyperparameter} &
			\multicolumn{1}{r}{Value} \\
			\midrule
			\multicolumn{1}{l}{Number of epochs} & $50$ \\
   			\multicolumn{1}{l}{Batch size} & $512$ \\
            \multicolumn{1}{l}{Negative log-likelihood} & $\text{Huber}_{\delta\mathbin{=}100}$~\cite{huber_loss} \\
            %\multicolumn{1}{l}{Huber loss delta} & $100$ \\
            \multicolumn{1}{l}{Optimizer} & Adam~\cite{adam}\\
			\multicolumn{1}{l}{Learning rate} & $0.01$ \\
   			\multicolumn{1}{l}{Learning rate decay epoch} & $40$ \\
			\multicolumn{1}{l}{Learning rate decay factor} & $0.1$ \\
   			\multicolumn{1}{l}{Late prediction correction factor} & $1$ \\

            \cmidrule{1-2}
            \multicolumn{2}{c}{\emph{Backpropagation}} \\
            \cmidrule{1-2}
            \multicolumn{1}{l}{Dropout~\cite{dropout} probability (drop)} & $0.2$ \\
            \multicolumn{1}{l}{Weight initialization} & Kaiming uniform~\cite{kaiming_init} \\

            \cmidrule{1-2}
            \multicolumn{2}{c}{\emph{Bayes by Backprop}} \\
            \cmidrule{1-2}
            \multicolumn{1}{l}{Prior} & $\mathcal{N}(\mathbf{0}, \mathds{I}\cdot 0.01)$ \\
            \multicolumn{1}{l}{Surrogate posterior} & $\mathcal{N}(\mathbf{0}, \mathds{I}\cdot \mathrm{softplus}^2(1)$ \\
            \multicolumn{1}{l}{Number of Monte Carlo samples} & $10$ \\

            \cmidrule{1-2}
            \multicolumn{2}{c}{\emph{Stein variational gradient descent}} \\
            \cmidrule{1-2}
            \multicolumn{1}{l}{Prior} & $\mathcal{N}(\mathbf{0}, \mathds{I}\cdot 0.01)$ \\
            \multicolumn{1}{l}{Number of particles} & $10$ \\
            \multicolumn{1}{l}{Kernel} & Radial basis function\tnotex{a_rbf} \\

			\bottomrule
		\end{tabular}
        \centering
        \begin{tablenotes}[flushleft]
         	\footnotesize
         	\setlength{\labelsep}{-0.5pt}
            \item[a] \label{a_rbf}
            Bandwidth is chosen using the median\nobreakdash-based heuristic by Liu et al.~\cite{liu_svgd}.
        \end{tablenotes}
	\end{threeparttable}
\end{table}

%% file: results.tex
\begin{table*}[hbt!]
\label{tab:accuracy_table}
    \setlength{\tabcolsep}{2.75pt}
    \footnotesize
	\centering 
	\begin{threeparttable}
		\caption{Mean and standard deviation (std) of the performance metrics on the rectified test set averaged over $10$ random seeds ($0$ -- $9$) of Dense3 (D3) and Conv2Pool2 (C2P2) trained via backpropagation (BP), Bayes by Backprop (BBB) and Stein variational gradient descent (SVGD). Metrics followed by ``\text{$\ast$}" are computed using the uncertainty\nobreakdash-based heuristic described in \cref{subsec:uncertainty_informed_predictions}. The best values for each model are highlighted in \textbf{bold}. The best overall values are \setlength{\fboxsep}{1pt}\fbox{framed}.}
		\label{tab:performance}
		\begin{tabular}{l@{\hskip 17pt}l@{\hskip 8pt}|rrrrrrrrr|rrrrrrrr}
			\toprule
			\multirow{2.5}{*}{Subset} &
			\multirow{2.5}{*}{Metric} &
			
			\multicolumn{2}{c}{\hspace{3.5pt}D3-BP} &&
			\multicolumn{2}{c}{\hspace{6pt}D3-BBB} &&
   			\multicolumn{2}{c}{\hspace{6pt}D3-SVGD} &&
			\multicolumn{2}{c}{\hspace{6pt}C2P2-BP} &&
			\multicolumn{2}{c}{\hspace{6pt}C2P2-BBB} &&
   			\multicolumn{2}{c}{\hspace{6pt}C2P2-SVGD} \\
			\cmidrule(l{6.5pt}r{2pt}){3-4} \cmidrule(l{8pt}r{2pt}){6-7} \cmidrule(l{8pt}r{2pt}){9-10}
			\cmidrule(l{8pt}r{2pt}){12-13} \cmidrule(l{8pt}r{2pt}){15-16} \cmidrule(l{8pt}r{2pt}){18-19}
			  && {mean} & {std} &&
			{mean} & {std} &&
			{mean} & {std} &&
			{mean} & {std} &&
			{mean} & {std} &&
			{mean} & {std} \\
			\midrule

			FD001
			\multirow{4}{*}   
            & RMSE & $14.25$ & $0.55$ && $14.33$ & $0.57$ && \setlength{\fboxsep}{1pt}\fbox{$\mathbf{13.17}$} & $0.14$ && $17.48$ & $0.06$ && $22.29$ & $0.61$ && $\mathbf{17.35}$ & $0.09$ \\ & MAE & $10.09$ & $0.37$ && $10.27$ & $0.28$ && \setlength{\fboxsep}{1pt}\fbox{$\mathbf{9.55}$} & $0.15$ && $\mathbf{12.94}$ & $0.08$ && $15.76$ & $0.51$ && $12.98$ & $0.06$ \\ & Score & $394$ & $43$ && $435$ & $94$ && \setlength{\fboxsep}{1pt}\fbox{$\mathbf{334}$} & $16$ && $677$ & $18$ && $\numprint{4882}$ & $763$ && $\mathbf{648}$ & $21$ \\

			\cmidrule(r{2pt}){2-19}
            & RMSE\textsuperscript{$\ast$} & $-$ & $-$ && $14.18$ & $0.38$ && \setlength{\fboxsep}{1pt}\fbox{$\mathbf{13.03}$} & $0.12$ && $-$ & $-$ && $21.62$ & $0.57$ && $\mathbf{17.31}$ & $0.09$ \\ & MAE\textsuperscript{$\ast$} & $-$ & $-$ && $10.14$ & $0.23$ && \setlength{\fboxsep}{1pt}\fbox{$\mathbf{9.36}$} & $0.13$ && $-$ & $-$ && $15.27$ & $0.42$ && $\mathbf{12.93}$ & $0.05$ \\ & Score\textsuperscript{$\ast$} & $-$ & $-$ && $369$ & $60$ && \setlength{\fboxsep}{1pt}\fbox{$\mathbf{318}$} & $13$ && $-$ & $-$ && $\numprint{3738}$ & $803$ && $\mathbf{639}$ & $20$ \\

            \midrule

			FD002
			\multirow{4}{*}            
            & RMSE & $38.92$ & $9.62$ && $19.23$ & $0.25$ && \setlength{\fboxsep}{1pt}\fbox{$\mathbf{18.27}$} & $0.40$ && $19.55$ & $0.23$ && $19.86$ & $0.32$ && $\mathbf{19.26}$ & $0.23$ \\ & MAE & $33.14$ & $9.11$ && $14.77$ & $0.21$ && \setlength{\fboxsep}{1pt}\fbox{$\mathbf{13.81}$} & $0.29$ && $15.43$ & $0.18$ && $15.43$ & $0.19$ && $\mathbf{15.17}$ & $0.18$ \\ & Score & $\numprint{60994}$ & $\numprint{29513}$ && $\numprint{2743}$ & $235$ && \setlength{\fboxsep}{1pt}\fbox{$\mathbf{\numprint{2259}}$} & $319$ && $\numprint{2699}$ & $136$ && $\numprint{3400}$ & $734$ && $\mathbf{\numprint{2470}}$ & $139$ \\

			\cmidrule(r{2pt}){2-19}
            & RMSE\textsuperscript{$\ast$} & $-$ & $-$ && $19.28$ & $0.24$ && \setlength{\fboxsep}{1pt}\fbox{$\mathbf{18.60}$} & $0.37$ && $-$ & $-$ && $19.85$ & $0.30$ && $\mathbf{19.24}$ & $0.23$ \\ & MAE\textsuperscript{$\ast$} & $-$ & $-$ && $14.82$ & $0.21$ && \setlength{\fboxsep}{1pt}\fbox{$\mathbf{14.07}$} & $0.27$ && $-$ & $-$ && $15.50$ & $0.19$ && $\mathbf{15.20}$ & $0.19$ \\ & Score\textsuperscript{$\ast$} & $-$ & $-$ && $\numprint{2430}$ & $181$ && \setlength{\fboxsep}{1pt}\fbox{$\mathbf{\numprint{2034}}$} & $268$ && $-$ & $-$ && $\numprint{3139}$ & $617$ && $\mathbf{\numprint{2356}}$ & $129$ \\

			\midrule
			
			FD003
			\multirow{4}{*}
            & RMSE & $14.97$ & $2.53$ && $15.97$ & $0.81$ && \setlength{\fboxsep}{1pt}\fbox{$\mathbf{12.33}$} & $0.24$ && $\mathbf{19.75}$ & $0.40$ && $22.79$ & $0.79$ && $19.99$ & $0.28$ \\ & MAE & $10.56$ & $2.15$ && $11.90$ & $0.51$ && \setlength{\fboxsep}{1pt}\fbox{$\mathbf{8.53}$} & $0.15$ && $\mathbf{14.60}$ & $0.33$ && $16.70$ & $0.51$ && $14.81$ & $0.16$ \\ & Score & $625$ & $407$ && $725$ & $198$ && \setlength{\fboxsep}{1pt}\fbox{$\mathbf{307}$} & $20$ && $\mathbf{\numprint{1430}}$ & $144$ && $\numprint{3233}$ & $590$ && $\numprint{1491}$ & $111$ \\

            \cmidrule(r{2pt}){2-19}
            & RMSE\textsuperscript{$\ast$} & $-$ & $-$ && $15.31$ & $0.46$ && \setlength{\fboxsep}{1pt}\fbox{$\mathbf{12.13}$} & $0.21$ && $-$ & $-$ && $21.95$ & $0.68$ && $\mathbf{19.88}$ & $0.28$ \\ & MAE\textsuperscript{$\ast$} & $-$ & $-$ && $11.53$ & $0.26$ && \setlength{\fboxsep}{1pt}\fbox{$\mathbf{8.45}$} & $0.14$ && $-$ & $-$ && $16.19$ & $0.44$ && $\mathbf{14.74}$ & $0.16$ \\ & Score\textsuperscript{$\ast$} & $-$ & $-$ && $561$ & $136$ && \setlength{\fboxsep}{1pt}\fbox{$\mathbf{287}$} & $17$ && $-$ & $-$ && $\numprint{2587}$ & $501$ && $\mathbf{\numprint{1447}}$ & $112$ \\

			\midrule

			FD004
			\multirow{4}{*}
            & RMSE & $36.20$ & $10.68$ && $21.77$ & $0.13$ && \setlength{\fboxsep}{1pt}\fbox{$\mathbf{21.00}$} & $0.31$ && $22.88$ & $0.24$ && $22.98$ & $0.43$ && $\mathbf{22.07}$ & $0.15$ \\ & MAE & $29.72$ & $10.04$ && $16.40$ & $0.11$ && \setlength{\fboxsep}{1pt}\fbox{$\mathbf{15.51}$} & $0.18$ && $17.66$ & $0.23$ && $17.52$ & $0.31$ && $\mathbf{16.71}$ & $0.22$ \\ & Score & $\numprint{83040}$ & $\numprint{61583}$ && \setlength{\fboxsep}{1pt}\fbox{$\mathbf{\numprint{6320}}$} & $732$ && $\numprint{6648}$ & $\numprint{1119}$ && $\numprint{7310}$ & $850$ && $\mathbf{\numprint{6537}}$ & $\numprint{1052}$ && $\numprint{6577}$ & $782$ \\

			\cmidrule(r{2pt}){2-19}
            & RMSE\textsuperscript{$\ast$} & $-$ & $-$ && $21.66$ & $0.12$ && \setlength{\fboxsep}{1pt}\fbox{$\mathbf{21.02}$} & $0.30$ && $-$ & $-$ && $22.87$ & $0.40$ && $\mathbf{22.00}$ & $0.16$ \\ & MAE\textsuperscript{$\ast$} & $-$ & $-$ && $16.29$ & $0.11$ && \setlength{\fboxsep}{1pt}\fbox{$\mathbf{15.49}$} & $0.21$ && $-$ & $-$ && $17.47$ & $0.30$ && $\mathbf{16.64}$ & $0.22$ \\ & Score\textsuperscript{$\ast$} & $-$ & $-$ && \setlength{\fboxsep}{1pt}\fbox{$\mathbf{\numprint{5657}}$} & $582$ && $\numprint{5830}$ & $\numprint{1061}$ && $-$ & $-$ && $\mathbf{\numprint{6050}}$ & $920$ && $\numprint{6141}$ & $687$ \\

			\bottomrule

		\end{tabular}
	\end{threeparttable}
\end{table*}